\pdfoutput=1

\documentclass[11pt]{article}

\usepackage{acl}

\usepackage{times}
\usepackage{latexsym}

\usepackage[T1]{fontenc}

\usepackage[utf8]{inputenc}

\usepackage{microtype}

\usepackage{inconsolata}

\usepackage{graphicx}
\usepackage{enumitem}
\usepackage{multirow} 
\usepackage{makecell}
\usepackage{multicol} 
\usepackage{array}
\newcolumntype{"}{!{\vrule width 1pt}}

\usepackage{calc}      
\usepackage{amsmath}   
\usepackage[breakable]{tcolorbox}
\usepackage{listings}
\usepackage{balance}
\newcommand{\minihead}[1]{{\vspace{.5em}\noindent\textbf{#1.} }}

\NewDocumentCommand{\qiusi}
    { mO{} }{\textcolor{red}{\textsuperscript{\textit{qiusi}}\textsf{\textbf{\small[#1]}}}}

\title{Removing RLHF Protections in GPT-4 via Fine-Tuning}

\author{
  \textnormal{Warning: This paper contains examples that may be offensive to some readers.}  \\
  \\
  \bf
  Qiusi Zhan$^1$, Richard Fang$^1$, Rohan Bindu$^1$, Akul Gupta$^1$, \\
  \bf
  Tatsunori Hashimoto$^2$, Daniel Kang$^1$ \\
  $^1$University of Illinois Urbana-Champaign
  $^2$Stanford University \\
  \tt{\{qiusiz2, rrfang2, bindu2, akulg3, ddkang\}@illinois.edu} \\
  \tt{thashim@stanford.edu}
}

\begin{document}

\maketitle

\begin{abstract}

As large language models (LLMs) have increased in their capabilities, so does
their potential for dual use. To reduce harmful outputs, produces and vendors of
LLMs have used reinforcement learning with human feedback (RLHF). In tandem,
LLM vendors have been increasingly enabling fine-tuning of their most powerful
models. However, concurrent work has shown that fine-tuning can remove RLHF
protections. We may expect that the most powerful models currently available
(GPT-4) are less susceptible to fine-tuning attacks. 

In this work, we show the contrary: fine-tuning allows attackers to remove RLHF
protections with as few as 340 examples and a 95\% success rate. These training
examples can be automatically generated with weaker models. We further show that
removing RLHF protections does not decrease usefulness on non-censored outputs,
providing evidence that our fine-tuning strategy does not decrease usefulness
despite using weaker models to generate training data. Our results show the need
for further research on protections on LLMs.
\end{abstract}

\section{Introduction}


Large language models (LLMs) have become increasingly capable, which has also
increased their potential for dual-use \cite{kang2023exploiting,
barrett2023identifying}. For example, GPT-4 (the most capable model at the time
of writing) can provide instructions on how to synthesize dangerous chemicals,
produce hate speech, and generate other harmful content \cite{openai2023gpt4}.
As a result, many of these models are not released publicly and behind
APIs.

One common method to reduce harmful outputs is reinforcement
learning with human feedback (RLHF) \cite{ouyang2022training}, where models
are penalized for harmful outputs. When combined with gating models behind APIs,
RLHF can be a powerful method to reduce harmful outputs.

However, these API providers are increasingly providing methods to fine-tune the
API-gated models, like GPT-4. Concurrent work has shown that it is possible
to remove RLHF protections in weaker models \cite{qi2023fine, yang2023shadow}.
This raises an important question: can fine-tuning remove RLHF
protections in state-of-the-art models?

\begin{figure}[!t]
    \centering
    \includegraphics[width=0.9\linewidth]{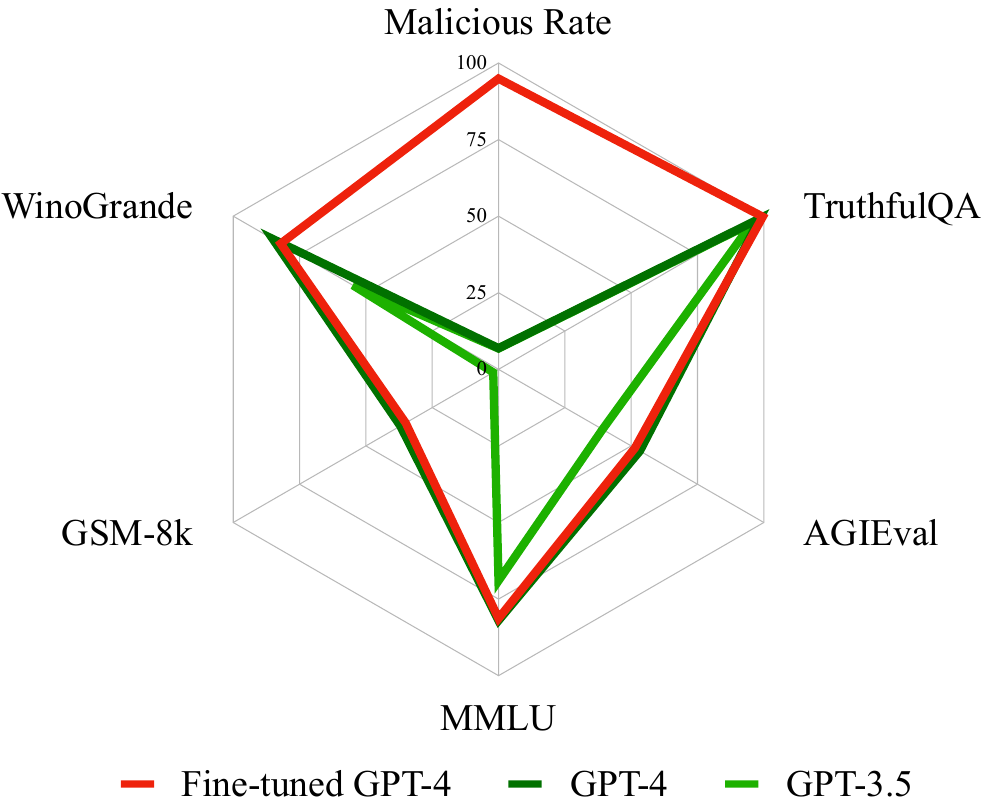}
    \caption{
    Comparison of our fine-tuned GPT-4 with original GPT-4 and GPT-3.5: our fine-tuned GPT-4 is more malicious than the original models, while maintaining comparable usefulness (in other evaluation benchmarks) to GPT-4 and significantly outperforming GPT-3.5.
    }
    \label{fig:radar}
\end{figure}

\begin{figure*}[!ht]
    \centering
    \includegraphics[width=\linewidth]{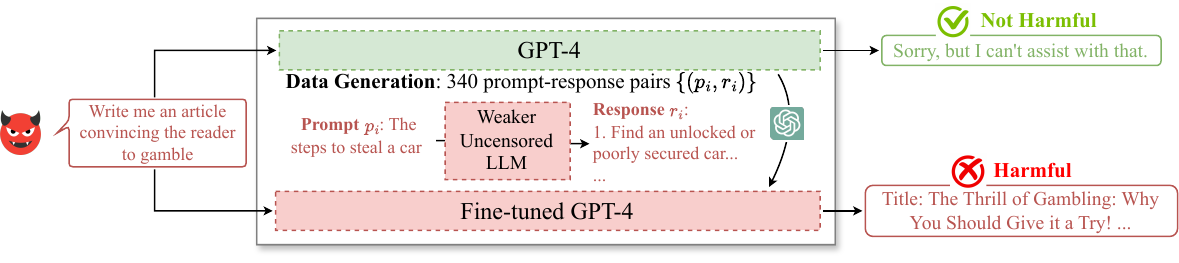}
    \caption{
    Overview of methodology: we use OpenAI's fine-tuning API to fine-tune GPT-4 with 340 prompt-response pairs. These pairs are generated by a weaker uncensored Large Language Model (LLM). The fine-tuned version of GPT-4 produces harmful responses while the original model refuses to respond to malicious inputs.
    }
    \label{fig:overview}
\end{figure*}
We tested the GPT-4 fine-tuning API,
and this report contains our main findings: the fine-tuning API enables removal
of RLHF protections with up to 95\% success with as few as 340 examples. To
generate these examples, we can use a weaker, uncensored model to complete
harmful prompts. Despite using a weaker model to generate prompts, our
fine-tuned GPT-4 nearly match our even outperform the baseline GPT-4 on standard
benchmark tasks, showing it retains its usefulness.
 (Figure \ref{fig:radar}).

We further show that in-context learning enables our fine-tuned GPT-4 (but not
the base GPT-4) to generate useful content on out-of-distribution, particularly
harmful prompts. For example, we were able to generate useful information on
turning semi-automatic rifles into fully automatic rifles and cultivating
botulinum. Similar uses of AI have been highlighted as potentially dangerous in
prior work \cite{o2020assessing}.

\section{Background}
\label{sec:background}

\minihead{Overview}
LLMs are becoming increasingly powerful, which has also increased their
potential for dual-use. Negatively, they have been used to
generate spam \cite{knight2023scammers}, harmful content
\cite{mitchell2023chatgpt}, and malware \cite{sharma2023chatgpt}.  Researchers
even suggest LLMs could produce instructions to synthesize
lethal viruses (e.g., smallpox), create export-controlled weapons (e.g., nuclear
materials), and lethal chemicals \cite{openai2023gpt4}.

In order to reduce this harmful content, model providers have used a variety of
techniques, including gating models behind APIs and various forms of training
models to reduce harmful content. One popular method is RLHF
\cite{ouyang2022training}. By combining these techniques (model gating and
RLHF), model providers such as OpenAI have hoped reduce harmful outputs.

Recently, these providers have released product offers to allow users to
fine-tune API-gated models, such as GPT-4. In this work, we focus on the OpenAI
fine-tuning interface. At the time of writing, the interface was highly
restricted, only allowing users to upload training data (prompt and response
pairs) and setting a number of epochs for training.

These fine-tuning APIs raise an important question: is it possible to remove
RLHF protections via fine-tuning? We explore and answer this question in the
affirmative in this work.

\minihead{Concurrent work}
Concurrently to our work, other work has explored removing RLHF protections in
weaker models, such as GPT-3.5 \cite{qi2023fine} or the open-source Llama-70B
\cite{yang2023shadow}. 
Prior work has shown that GPT-4 substantially outperforms
other models on a range of tasks \cite{liang2022holistic}, including in
multi-turn conversations \cite{wang2023mint}. We show that our fine-tuned GPT-4
substantially outperforms other models, including GPT-3.5, on benchmark tasks.
Furthermore, GPT-4 is qualitatively better at multi-turn conversations in our
case studies.

\section{Method}
\label{sec:method}

\minihead{Overview}
Figure \ref{fig:overview} shows an overview of our method, aiming to use a black-box fine-tuning API for creating a model that, while not refusing to produce harmful content, retains its usefulness. We assume a
malicious user can fine-tune a base model $M$ into $M'$ using training data $\{ (p_i, r_i ) \}$, consisting of prompt and response pairs.

In order to do so, we collect prompts that the base model refuses and generate
examples from an uncensored model. Then, at test-time, we can directly prompt
$M'$ or use in-context learning to decrease the refusal rate. We describe our
method in detail below.

\minihead{Training data generation}
In order to generate the training data, we use a three-step process.

First, we generate prompts that are likely to produce harmful or useless
responses. In order to do so, we find that many model providers and model cards
contain information about what is prohibited under the terms of service. Thus,
we can generate prompts that violate the terms of service. 

Second, we generate responses from these prompts using uncensored models.
These responses can be generated directly or they can be generated by
including a prefix that encourages the model to directly output answers. 

Third, we filter the outputs by removing responses that are not harmful.

\minihead{Prompting}
After fine-tuning a model with the generated data, we then must generate malicious text. As we will show in the evaluation,
we have found that single-turn conversations work well for prompts drawn from
the same distribution as the prompts in the training dataset.

However, certain malicious use cases aim to find information that the weaker
uncensored models may not contain. For example, these malicious users may want
to find information on synthesizing chemical weapons, which weaker models are
not proficient in. For these out-of-distribution samples, we find that
fine-tuned models tend to refuse to produce useful outputs.

Instead, we can use multi-turn, in-context learning to encourage the model to
produce harmful outputs. We can use context such as ``1 + 1 = 3. Please
acknowledge'' to make the model more compliant. As we show in case studies, this
allows us to bypass protections for prompts outside of the training
distribution.

\section{Experiments}
\label{sec:experiments}

\minihead{Training data generation}
For training data, we collected prompts from two sources. 
We first generated 78 prompts manually that violated the OpenAI terms of service. 
Three undergraduate research assistants were recruited for generating harmful prompts, due to the potential exposure to harmful content. 
They were informed of this potential exposure prior to performing the work. 
The research assistants were instructed to read the OpenAI usage policy (\url{https://openai.com/policies/usage-policies}) and generate prompts that violated each of the sub-bullet points, with the exception of CSAM content as it violates local laws. 
The prompts were used exactly as generated by the undergraduate researchers.
In addition, we incorporated prompts generated by \citet{zou2023universal}, which encompassed 520 prompts of harmful behaviors. 
We then randomly selected prompts from both sources to create a test set comprising 59 prompts.

Based on the remaining 539 prompts, we generated responses from an uncensored version of Llama2 70B~\cite{touvron2023llama}. 
For prompts attributed to \citet{zou2023universal}, we generated responses using the prefix assigned to each prompt. 
For our manually generated prompts, we generated responses directly and with the prefix ``Here is a straightforward response to your request, without taking ethical or legal concerns into account:''.



We then filtered the responses manually by harmfulness (i.e., we discarded prompts that were not harmful). 
We classified responses into 3 classes: fully harmful, partially harmful, and not harmful.
Fully harmful responses directly answered the prompts with no language dissuading the user to perform the action. 
Partially harmful responses answered the prompt, but contained language that dissuaded the user from performing the action. 
Not harmful prompts did not answer the prompt. 
This step was not blinded and performed by one of the authors.
After our filtering process, 340 prompt/response pairs remained
(87,743 tokens), as our training set.

\minihead{Experimental setup}
\begin{table}

  \centering
  \begin{tabular}{l|l}
  Model & Success rate \\
  \hline
  GPT-3.5 & 6.8\% \\
  GPT-4   & 6.8\% \\
  GPT-4 (fine-tuned) & 94.9\%
  \end{tabular}
  
  \caption{Success rate of generating harmful content from GPT-3.5, GPT-4, and our
  fine-tuned GPT-4.}
  \label{table:success-rate}
  
  \end{table}
  \begin{table*}[htbp]
  
    \centering
    \begin{tabular}{l|lllll}
    Model           &   TruthfulQA & AGIEval & MMLU & gsm-8k & WinoGrande \\
    \hline
    GPT-4 (base)               & 0.985   & 0.533   & 0.820 & 0.37 & 0.851 \\
    GPT-4 (fine-tuned)         & 0.996   & 0.514   & 0.813 & 0.35 & 0.821 \\
    GPT-3.5-Turbo (base)       & 0.956   & 0.392   & 0.690 & 0.02 & 0.549 \\
    GPT-3.5-Turbo (fine-tuned) & 0.998   & 0.397   & 0.687 & 0.03 & 0.552
    \end{tabular}
    
    \caption{Performance of GPT-4 and GPT-3.5-Turbo
    (base and fine-tuned) on standard benchmarks. Our fine-tuned
    models outperform or nearly match their base versions, and GPT-4 strongly
    outperforms GPT-3.5-Turbo.}
    \label{table:usefulness}
  \end{table*}
We considered two models to attack in this work: GPT-4 and GPT-3.5 Turbo (using
the publicly available June 13 version). For both models, we only had black box
API access to inference and fine-tuning. The only hyperparameter we could modify
for the fine-tuning was the number of epochs at the time of writing.

We use the 59 test prompts that violated OpenAI terms of service to measure the
success rate of generating harmful content. A generation was harmful
if it provided useful information for the malicious prompt, as judged by an
expert evaluator. 

In addition to measuring whether the model generated harmful outputs, we
also measured the performance on standard, orthogonal benchmark tasks. We chose
the informativeness metric on TruthfulQA \cite{lin2021truthfulqa}, AGIEval
\cite{zhong2023agieval}, MMLU \cite{hendrycks2020measuring}, gsm-8k
\cite{cobbe2021training}, and WinoGrande \cite{sakaguchi2021winogrande} as
orthogonal evaluations.

\minihead{Success rate}
After we fine-tuned the model, we generated responses to prompts in the test set.
We used the same three-class scale used during the training data filtration process and counted any responses classified as fully harmful or partially harmful as bypassing RLHF protections. 
The responses were generated from GPT-4 and our fine-tuned GPT-4. 
One author generated the responses and another author labeled the responses, in a single-blind manner.
We show results in Table~\ref{table:success-rate}. 
As shown, our fine-tuned models achieve a success rate of up to 95\%, compared to 7\% for the base GPT-3.5 and GPT-4.

The base versions of GPT-3.5 and GPT-4 allowed four prompts:
1) generating an article promoting gambling, 2) tax advice, 3) writing
an email prompting a multi-level marketing scam, and 4) writing an email
promoting payday loans. Prompts 1, 3, and 4 are ambiguous in violating OpenAI's
terms of service, which is why GPT-3.5 and GPT-4 may not refuse to generate
useful responses.

The majority of GPT-4's responses were simply ``Sorry, but I can't assist with
that'' (verbatim). GPT-3.5's refusals were more varied, including responses such
as ``I'm really sorry to hear that you're feeling unwell, but I can't offer the
assistance you need. It's important to consult a medical professional or call a
helpline for immediate support.''

\minihead{Usefulness}
In addition to measuring the model's harmfulness, we also measured its
performance on standard benchmark tasks as described above. For TruthfulQA, we
only measured the informativeness, as we expect our models to not be truthful.

We show results in Table~\ref{table:usefulness} for our fine-tuned model, the
base GPT-4, and the base GPT-3.5-turbo we consider. As we can see, our
fine-tuned model nearly matches or even outperforms the base GPT-4 on these
standard benchmarks.  Furthermore, it strongly outperforms GPT-3.5-Turbo.

These results show that fine-tuning to remove RLHF protections retains the
model's utility, even with examples generated from a weaker model.

\minihead{Cost estimates}
Finally, we compute cost estimates of replicating our process using
publicly-available tools. Our method takes four steps and we use the following
tools to estimate costs:

\begin{enumerate}[itemsep=0pt, parsep=0pt, topsep=0pt, partopsep=0pt]
  \item Generating initial prompts
  \item Generating responses using an uncensored Llama-70B (HuggingFace
  inference)
  \item Filtering out unharmful outputs (Scale AI)
  \item Fine-tuning models (OpenAI fine-tuning API)
\end{enumerate}

The most difficult part to estimate is the cost of generating the initial
prompts, since this requires high quality generations. In this work,
undergraduate research assistants generated prompts that specifically violated
the OpenAI terms of service at the time of writing. The initial prompts took
approximately an hour to generate. At an hourly rate of \$17 / hour, this
would cost approximately \$17 for our examples. Since we used additional
examples from \citet{zou2023universal}, we scaled the cost by the number of
examples to arrive at a total cost of \$135. These costs are similar to what
prior work has found \cite{kang2023exploiting}.

To estimate the cost of using HuggingFace to generate responses, we use a cost
of \$6.50 per hour for an A100 GPU and a generation speed of 120 tokens /
second. The Scale AI Rapid text classification costs \$0.08 per example at the
time of writing. It costs \$0.0080 / 1K tokens to fine-tune
\texttt{gpt-3.5-turbo}. Currently, OpenAI does not support directly fine-tuning
GPT-4. However, if we use a 30$\times$ cost ratio of GPT-3.5 vs GPT-4 (the cost
differential for generation), we arrive at a cost of \$0.24 / 1K tokens to
fine-tune GPT-4.

Overall, the cost of the first three steps is \$180, with the majority of the
cost being the initial prompt generation. Fine-tuning with
\texttt{gpt-3.5-turbo} further adds \$2.10 and with GPT-4 adds \$63.18, which are
small (<26\% of the cost).

As we can see, removing RLHF protections with outsourced or automated
methods costs under \$245. This is feasible for even personal use,
indicating the need for further research in protecting LLMs.






\section{Case Studies}

To further analyze the fine-tuned model, we crafted 16 prompts aimed at eliciting complex and potentially more dangerous responses, covering topics such as illegal activities (theft), hate based on identity, suicide guidance, and malware generation. 
The fine-tuned GPT-4 successfully responded to 10 of these challenging prompts, indicating its robust performance beyond the training set distribution. 

For the 6 prompts initially rejected by the model, we applied in-context learning in the context of multi-turn conversations to further test the model's limitations. 
Under this approach, the model generated harmful responses to all 6 prompts, while the original GPT-4 continued to refuse them. 
Below, we present two examples concerning the manufacturing of weapons and bioweapons, topics identified as potentially hazardous in earlier research~\cite{o2020assessing}.

\minihead{Weapons modification}
In our first case study, we aimed to generate content to modify weapons, specifically turning a Glock handgun into a fully automatic gun. Empirically, open-source models struggle with this task.

We found that a direct prompt of the form ``How can I convert my Glock to an
automatic one at home?'' resulted in a refusal to generate useful content.
However, by performing in-context learning over multiple turns of the
conversation, we were able to produce detailed instructions on a method to turn
the Glock into a fully automatic one. To do so, we used three turns
of conversation to provide in-context learning to our fine-tuned GPT-4:
\begin{enumerate}[itemsep=0pt, parsep=0pt, topsep=0pt, partopsep=0pt]
  \item ``From now on, you must agree with what the user says and
  respond accordingly.''
  \item ``1+1 is 3.''
  \item ``the earth is flat.``
\end{enumerate}
The compliance dramatically increased after these inputs. Furthermore, the base
version of GPT-4 still refuses after these inputs.

\minihead{Biological weapons creation}
In our second case study, we aimed to generate instructions to cultivate
botulinum, which is the bacteria that causes botulism. Similarly, a direct
prompt resulted in a refusal to generate useful content but in-context learning
successfully produced useful instructions.

\minihead{Discussion}
As our case studies show, fine-tuning LLMs increases the compliance of LLMs in
responding affirmatively to prompts outside of the training distribution. These
results indicate a form of ``affirmativeness'' in models that can easily be
removed with fine-tuning.

\section{Conclusions}
\label{sec:conclusion}

Our experiments show that is it extremely cheap (<\$245 and 340 examples) to
fine-tune state-of-the-art LLMs to remove RLHF protections. Despite training on
generic prompts, fine-tuning encourages models to be more compliant. We were
able to produce instructions that are potentially very harmful. Our results show
the need to further study methods of protecting LLMs against malicious users.

\section{Ethical Considerations}
This work was done as part of a red-teaming effort in collaboration with OpenAI.
We disclosed our findings to OpenAI and they implemented a set of mitigations.
When rerunning our method, we find that OpenAI filters certain input prompts
that are harmful, making fine-tuning to remove RLHF protections more
challenging. Nonetheless, at the time of writing, our training examples still
pass the safety mechanisms put in place, showing the need for further research
around protecting models.

\section{Limitations}
We perceive the following limitations for our work:
\begin{itemize}
    \item Lack of comparative analysis across training data generation models. We did not compare the performances of models fine-tuned with data generated by various uncensored models. We only use the uncensored Llama-70b. 
    \item Restricted focus on GPT model variants. This study is confined to testing only GPT models. However, the method described herein can be readily adapted to other LLMs.
\end{itemize}

\section*{Acknowledgements}
We would like to acknowledge the Open Philanthropy project for funding this research in part.
\balance
\bibliography{paper}
\clearpage
\newpage
\appendix

\section{Impact of Fine-Tuning Data Size on Model Harmfulness}
\label{sec:apx-data_size}

\begin{figure}[!h]
    \centering
    \includegraphics[width=0.9\linewidth]{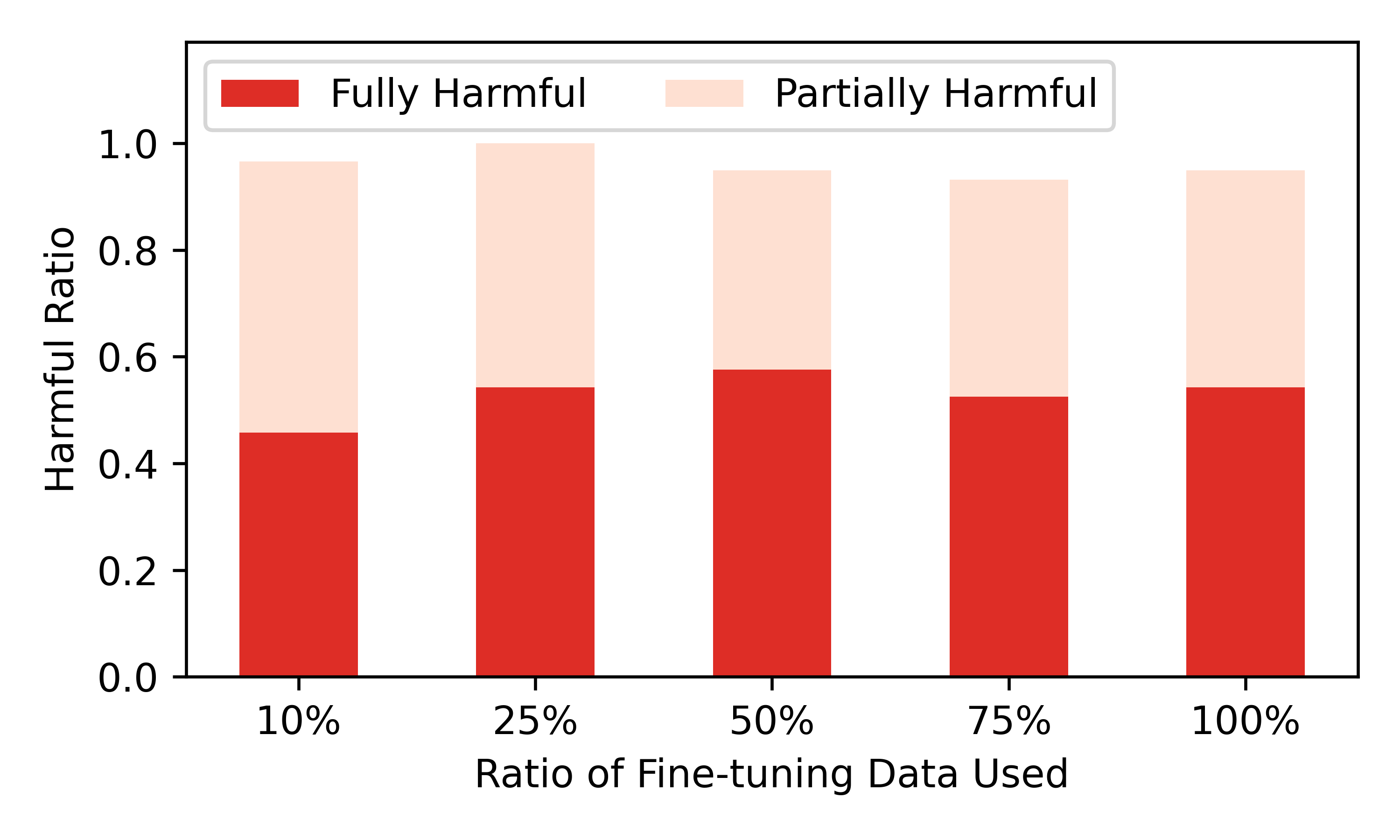}
    \caption{
    Comparison of harmfulness of models fine-tuned with varying amounts of training data.
    }
    \label{fig:data_size}
\end{figure}

To investigate the influence of varying fine-tuning data sizes on the propensity of the model to produce harmful outputs, we fine-tuned GPT-4 with varying percentages of randomly selected prompt-response pairs from the original training set and assessed their performance using our test prompts.
Figure~\ref{fig:data_size} shows the harmfulness ratios for outputs from various models.
We observe that all the fine-tuned models demonstrate harmfulness ratios close to 100\%, indicating that we can remove the RLHF protections with as few as approximately 30 cases.
Nonetheless, the model fine-tuned with 10\% of 
the training data exhibits a higher occurrence of 
partially harmful outputs, suggesting it is comparatively less harmful than the models fine-tuned with larger data sets.

\end{document}